\documentclass{IEEEtran}
\usepackage{amsmath,amssymb,amsfonts}
\usepackage{algorithmic}
\usepackage{graphicx}
\usepackage{textcomp}
\usepackage[numbers]{natbib}
\usepackage{hyperref}
\usepackage{tikz}
\usepackage{makecell}
\usetikzlibrary{positioning, backgrounds, calc}

\def\BibTeX{{\rm B\kern-.05em{\sc i\kern-.025em b}\kern-.08em
    T\kern-.1667em\lower.7ex\hbox{E}\kern-.125emX}}

\begin{document}

\title{DAEDRA: A language model for predicting outcomes in passive pharmacovigilance reporting}

\author{Chris von Csefalvay, \IEEEmembership{Member, IEEE}

\thanks{Submitted on XX February, 2024. This work was supported, in part, by intramural research funding from HCLTech (to CvC). The funders had no role in study design, data collection and analysis, decision to publish, or preparation of the manuscript.}
\thanks{Chris von Csefalvay is with Enterprise and Research Services, HCLTech, Vienna, VA 22182 USA (e-mail: \texttt{chris.csefalvay@ieee.org}).}}

\markboth{IEEE Journal of Translational Engineering in Health and Medicine,~Vol.~X, No.~X, X~2024}%
{von Csefalvay: DAEDRA: A language model for predicting outcomes in passive pharmacovigilance reporting}

\maketitle

\begin{abstract}
Over the recent years, the emergence of large language models (LLMs) has given rise to a proliferation of domain-specific models that are intended to reflect the particularities of linguistic context and content as a correlate of the originating domain.
This paper details the conception, design, training and evaluation of DAEDRA, a LLM designed to detect regulatory-relevant outcomes (mortality, ER attendance and hospitalisation) in adverse event reports elicited through passive reporting (PR).
While PR is a highly cost-efficient way of eliciting information from a wide and diverse audience -- typically including not only physicians and healthcare providers but also patients, family members and other lay stakeholders --, this diversity makes PR corpora difficult to analyse.
Generic language models may not capture the complex clinical dimensions while specific clinical or biomedical models may not perform well on lay reports.
To evaluate the utility of a subdomain-specific language model, an adaptive training approach was adapted, wherein base language model candidates were evaluated on a subset of the corpus, and the best performer was trained on the entire corpus.
This yielded a small but significant improvement in $F_1$ (+1\%), precision (+2.5\%) and recall (+3.8\%), at a relatively low training cost and a single-day training time.
Subdomain-specific LLMs continue to be viable options for better results when analysing highly specialised corpora.
\end{abstract}


\section{Introduction}
\label{sec:introduction}

In passive pharmacovigilance reporting (PR), also referred to as spontaneous reporting, reports of adverse drug reactions (ADRs) are not directly elicited but rather supplied by patients, clinicians and sometimes even entirely unrelated parties.
Such reporting is typically voluntary where patients are concerned, but may place mandatory reporting obligations on clinicians.
For instance, in the United States, there exists a mandatory reporting obligation for certain vaccines administered in childhood, pursuant to the National Childhood Vaccine Injury Act of 1986 (42 U.S.C. §300aa-1 to 300aa-34), and for certain events pertaining to SARS-CoV-2 vaccines, under their respective Emergency Use Authorisations or Biologics License Applications. 
PR is an efficient and highly cost-effective method of eliciting information about potential ADRs, as it does not require the costly and complex follow-up common in passive pharmacovigilance reporting. 
This makes PR eminently suitable for scenarios where the number of expected ADRs in relation to the number of exposed individuals is low.
Several highly successful PR schemes are in widespread use, and the availability of online reporting has only augmented the power of these systems. 
Some of the best known are the UK's Yellow Card Scheme\cite{ODonovan2022IdentifyingScheme} and the Vaccine Adverse Event Reporting System (VAERS)\cite{Shimabukuro2015SafetyVAERS}, maintained by the FDA.

The principal drawback of PR is, fundamentally, one of data quality and governance.
In particular, much of the payload of such reporting lies in unstructured (plain text) fields, which makes structured analysis quite difficult.\cite{ODonovan2022IdentifyingScheme}
Until recently, such analysis required manual coding of reports or reliance on the reporter's completion of a form that would often be complex or ambiguous for a layperson -- for instance, a layperson may not know to differentiate between emergency room (ED) attendance \textit{versus} hospital admission.
Prior work in accessing the unstructured (plain text) element of VAERS reports has been limited both in its scope and in its success. 
\citet{Botsis2011TextSelection} describe a conventional text mining approach using a rule-based classifier in conjunction with boosted trees and weighted support vector machines, but this work was validated only on a very small subset of reports.
\citet{Baer2016CanReview} used text mining to simplify reports and facilitate manual review, but not direct classification towards outcomes of interest.

The rise of novel language models has enabled the structured mining of such unstructured data assets, with Large Language Models (LLMs) -- a diffuse term encompassing a wide range of statistical models intended to analyse and evaluate unstructured text -- playing the most significant part.
Such models have been applied with considerable success to clinical text. 
Initial approaches, such as \citet{Liu2021ClinicalBERT} (in the context of clinical trials) and \citet{Turchin2023ComparisonDocuments} (in three clinical record contexts: bariatric surgery, statin therapy adherence and documentation of tobacco use status) have used generic BERT models (\citet{Devlin2019BERT:Understanding}) trained on domain-nonspecific data.
Early on, however, it was found that domain-specific training is essential in the biomedical context.
The comparison by \citet{Turchin2023ComparisonDocuments} was followed by the recent work of \citet{Scaboro2023ExtensiveExtraction}, whose findings conclusively show the superiority of domain-specific models in the biomedical realm. 

This paper reports on the development of DAEDRA (Determining Adverse Event Dispositions for Regulatory Affairs), which is to the best of the authors' knowledge the first LLM based approach specifically trained on a pharmacovigilance data set from PR, and the first LLM specifically designed to estimate severity outcomes from unstructured symptom descriptions and narratives.
The model is trained on a real-world pharmacovigilance data set spanning over three decades (1990-2023) and comprising over 1.8m records for a total corpus of 173,093,850 words constructed from a subset of reports submitted to VAERS.
The model is intended to identify, based on the narrative, whether any, or any combination, of three serious outcomes -- death, hospitalisation and ER attendance -- have occurred. 
DAEDRA is novel in three respects.

\begin{enumerate}
    \item It is the first model specifically trained on pharmacovigilance data from passive reporting. While clinical language models are gradually becoming ubiquitous, few of these target the specific field of pharmacovigilance. A strong case could be made that because pharmacovigilance tasks involve reasoning about unintended consequences of pharmacological intervention, such data is by definition out of sample to typical biomedical language models, which could be more accurately considered "business as usual" \textit{vis-a-vis} the way our examination of ADR reports probes the extrema of clinical practice.
    \item Most biomedical LLMs were trained either on scholarly literature or, in the case of (the very few and far between) clinical models, on a purely clinical underlying dataset. While this may have included the voices of diverse authors from diverse parts of the clinical care spectrum, the contents of these originate from providers/clinicians. This does not comport entirely with the typical corpus of passive reporting, where clinicians and laypersons alike are invited to participate.
    \item As a specific model of clinical outcomes based on an unstructured description of patient presentation, DAEDRA is a model that was trained with the inherent 'noisiness' of passive reporting in mind. This includes both linguistic noise (misspellings, colloquialisms, colloquial misuse of terminology) and content noise (irrelevant comments, subjective personal descriptions, political invective and at least one report of a dead spider found in the examination room).
\end{enumerate}

The motivation behind DAEDRA is twofold.
From a research perspective, it seeks to further test the hypothesis that subdomain-driven models trained on highly specific language are a viable approach towards improving the predictive power of base models over contexts that use domain-specific language.
From a translational perspective, it is intended to be a lightweight yet performant model for the quantification of ADRs of high regulatory relevance (mortality and the need for emergency medical attention or in-patient care) from PR corpora.

\section{Methods and materials}
\label{sec:methods}

\subsection{Source data}
\label{sub:source_data}

A comprehensive corpus of all reports made to VAERS, from 1990 to 2023, inclusive, was obtained. Due to the truncation of free-text report fields for certain non-US reports owing to regulatory limitations by European authorities, the decision was made to limit the scope of the corpus to reports made in respect of vaccination events in the United States and its territories. Reports were processed using \texttt{pandas} 2.0.2 in Python 3.8. 

Together with the relevant events, the unstructured \texttt{SYMPTOM\_TEXT} field was included in the data set, as well as the VAERS ID for later cross-correlation with other factors. Filtering to exclude fields without a valid \texttt{SYMPTOM\_TEXT} yielded a final tally of 1,814,920 records, with the text corpus spanning 173,093,850 words in total.

The following three relevant events were selected as target outcomes:

\begin{itemize}
    \item \textbf{mortality}: values of \texttt{Y} for VAERS variable \texttt{DIED};
    \item \textbf{ER attendance}: values of \texttt{Y} for either of VAERS variables \texttt{ER\_VISIT} or \texttt{ER\_ED\_VISIT} (to accommodate different versions of the VAERS form that use different terminology for the same event); and
    \item \textbf{hospitalisation}: values of \texttt{Y} for VAERS variable \texttt{HOSPITAL}.
\end{itemize}

Since the research objective is to identify constellations of presentations rather than each outcome on its own, a powerset encoding was used to represent each combination (e.g. ER attendance followed by mortality), as well as the absence of any relevant event, as a distinct class. 
This is also more computationally convenient as well as statistically robust, especially owing to the strong class imbalance.
77.2\% of records were associated with no target outcomes, with the remainder being dominated by ER attendance alone (14.8\% of all records), hospitalisation alone (3.9\% of all records) and ER attendance followed by hospitalisation (\% of all records). 
Only \% of all records involved a mortality outcome.

Records were subdivided into a training, testing and validation data set at split ratios of 70/15/15.
Record subdivision was performed by matching on gender (VAERS \texttt{SEX} field) and age quintile (quintile of VAERS \texttt{AGE\_YRS} field) to ensure that the splits reflected age and gender distributions of the original sample, as these are strongly correlated with the incidence of ADRs.

\subsection{Base model selection}

To select the most suitable base model, nine base models -- including seven domain-specific models and two domain-nonspecific models -- were trained on 10\% sample of the training set (7\% of the entire data set), with the $F_1$ score as the principal metric for selecting the most suitable base model. 
For base model selection, models were run on the same reference architecture as used for final training (described in Section~\ref{sub:model_training}), at a batch size of 64 for 3 epochs.
The models generally fell into three distinct categories:

\begin{enumerate}
    \item \textbf{General} models: these were base models that were originally trained on general corpora. BERT, for instance, was trained on a large general corpus consisting of the corpus of the English Wikipedia along with the Toronto BookCorpus, built from 7,000 self-published literary works.\cite{Devlin2019BERT:Understanding} This makes it a good proxy for casual language use, but as the weak performance of such models when compared to more specialised counterparts attests, not a good representation of clinical language.
    \item \textbf{Scientific} and \textbf{medical} models: these base models comprise the majority of the models under examination, and consist of models that were trained on domain-specific corpora. SciBert\cite{Beltagy2019SCIBERT:Text} was trained on a sample of papers from Semantic Scholar, comprising papers on computer science and the wider biomedical domain at a roughly 1:4 ratio. BioRoBERTa\cite{Gururangan2020DontTasks} also leverages Semantic Scholar, but is limited to papers from the biomedical domain. Other models used publicly available corpora of scientific literature. BioMedBERT\cite{Gu2022Domain-SpecificProcessing}, BioELECTRA\cite{Kanakarajan2021BioELECTRA:PretrainedDiscriminators}, TinyPubMedBERT\cite{Yoon2022BiomedicalFramework} predominantly used information from PubMed, a very large database of abstracts of papers from the biomedical domain maintained by the US National Institutes of Health's National Library of Medicine (NLM). BioBERT\cite{Lee2020BioBERT:Mining} stands out through the diversity of corpora, including not only PubMed but also PubMed Central (PMC), which incorporates a large corpus of full-text articles.
    \item \textbf{Clinical} models: unlike medical models, clinical models were trained on corpora produced in the ordinary course of clinical care, such as EMR/EHR records and patient notes. Because such records are generally hard to obtain due to issues of privacy and patient confidentiality, there is a relative dearth of such models, with \citet{Alsentzer2019PubliclyEmbeddings} disclosing BioClinicalBERT as perhaps one of the only genuine clinical language models in public availability. This model was trained on MIMIC-III,\cite{Johnson2016MIMIC-IIIDatabase} a large database of clinical notes from critical care.
\end{enumerate}

To facilitate comparison, the $F_1$ score was calculated.
This measure commends itself by two principal features. 
First, the $F_1$ score is relatively robust \textit{vis-a-vis} a significantly imbalanced data set, as is the case here (see Section~\ref{sub:source_data}).
In addition to this benefit, the $F_1$ score offers a convenient single measure that integrates both precision and accuracy.
It is calculated as the harmonic mean of precision and recall, or

$$ F_1 = \frac{2 \ TP}{2 \ TP + FP + FN} $$

\noindent where $TP$ and $FP$ denotes true positives and false positives, respectively, while $FN$ denotes false negatives.

The results of the comparison are laid out in Table~\ref{tab:comparison}. 
Of the models being compared, all domain-specific models except TinyPubMedBERT outperformed the generic models (the original BERT model and DistilBERT, a downscaled version of BERT).
It is notable that of the models, SciBERT -- which is a scientific language model but not domain-specific to medicine -- has performed quite competitively with the specifically medical language models, yielding a preferable $F_1$ score.

Performance among the leading models was quite similar, and either of the runner-ups (BioMedBERT and SciBERT) would have made suitable starting points.
Ultimately, the choice was made to evolve the BioBERT model (\texttt{biobert-base-cased-1.2}) as it performed slightly faster than its closest competitors.
A particularly gratifying finding of this comparison was that at least on our specific pharmacovigilance-driven data set, medical models -- which were domain-specific as to subject matter but not necessarily reflective of the linguistic context, whcih is closer to clinical care than biomedical scholarship -- were not inferior, and were for the most part superior, to a clinical-specific model (i.e. BioClinicalBERT\cite{Alsentzer2019PubliclyEmbeddings}).
Given the highly limited availability of such models and the considerable difficulty of obtaining corpora on which such models could be trained, the ability of domain-specific but not context-specific models to accommodate 'code switching' to the domain of clinical application is encouraging.

\begin{table*}
\renewcommand{\arraystretch}{1.3}
\caption{A comparison of candidate models.}
\label{tab:comparison}
\centering
\begin{tabular}{l|l|cccc}
\hline
\bfseries Model & \bfseries Domain & \bfseries Precision & \bfseries Recall & \bfseries $F_1$ & \bfseries Runtime (\textit{s}) \\
\hline\hline
SciBert\cite{Beltagy2019SCIBERT:Text} & Scientific & \textbf{0.7158} & 0.6325 & \textbf{0.8709} & 11,063.13 \\
BioBERT\cite{Lee2020BioBERT:Mining} & Medical & 0.7180 & \textbf{0.6920} & 0.8706 & 11,050.53 \\
BioMedBERT\cite{Gu2022Domain-SpecificProcessing} & Medical & 0.6958 & 0.6279 & 0.8706 & 11,176.73 \\
BioRoBERTa\cite{Gururangan2020DontTasks} & Medical & 0.7006 & 0.6354 & 0.8689 & 11,086.35 \\
BioClinicalBERT\cite{Alsentzer2019PubliclyEmbeddings} & Clinical & 0.7006 & 0.6354 & 0.8689 & 11,041.75 \\
BioELECTRA\cite{Kanakarajan2021BioELECTRA:PretrainedDiscriminators} & Medical & 0.6619 & 0.5791 & 0.8688 & 11,023.08 \\
BERT & General & 0.6871 & 0.5618 & 0.8603 & 3,878.07 \\
DistilBERT & General & 0.5406 & 0.4699 & 0.8579 & \textbf{2,129.35} \\
TinyPubMedBERT\cite{Yoon2022BiomedicalFramework} & Medical & 0.3765 & 0.2943 & 0.8421 & 2,380.33 \\
\hline
\end{tabular}
\end{table*}

\subsection{Tokeniser training}

Language models operate on sequences of integer representations of individual tokens, which may be words or fragments of words.
Tokenisers convert words into tokens or sequences of tokens according to learned rules, based on a vocabulary. 
It stands to reason that a model's performance is fundamentally constrained by the accuracy of the tokenisation: incorrect tokens deprive the model of the ability to meaningfully interpret terms. 

In the medical domain, especially where the task pertains to predicting the severity of a clinical outcome, the absence of specific tokens may lead to the loss of crucial meaning.
Consider, for instance, the term 'intussusception', a rare but serious complication of the rotavirus vaccine that frequently results in hospitalisation. 
A naive tokeniser would tokenise it as \texttt{int, \#\#uss, \#\#us, \#\#ception} (where \texttt{\#\#} denotes subword tokenisation, to differentiate it from a free-standing instance of the same morpheme). 
Neither of these tokens are either particularly unique nor particularly meaningful to connect to a hospitalisation event as opposed to other outcomes. 
On the other hand, a tokeniser specifically trained on the corpus at hand will consider \texttt{intussusception} as a free-standing token of its own, and one that is highly correlated with a hospitalisation outcome.
Other concepts that a domain-specific tokeniser is more likely to accommodate include 

\begin{itemize}
    \item \textbf{proper nouns}, e.g. \texttt{prevnar}, \texttt{pfizer};
    \item \textbf{abbreviations}, e.g. \texttt{JEE}, \texttt{POTS}; and
    \item \textbf{shorthand}, e.g. \texttt{y/o} (for 'years old'), \texttt{S/P} (for 'status post').
\end{itemize}

To accommodate the special linguistic features of the corpus, the base model's tokeniser was re-trained on the training set's corpus, and the vocabulary size was expanded from 28,996 to 52,000.
Re-tokenisation uses the same WordPiece tokeniser used by the original model, which is implemented using the \texttt{transformers} library's \texttt{BertTokenizerFast} class.
A principal benefit of WordPiece is that it reduces the problems posed by out-of-vocabulary (OOV) words through splitting words into subword tokens.
As such, WordPiece is a good choice for specialised language, but at the cost of doing so by relying on a greedy algorithm that may choose a more efficient but less valid tokenisation.
This is a particular issue for medical terminology, where compound loanwords are quite prevalent.
It would, thus, tokenise \texttt{pancreatitis} into \texttt{pan, \#\#cre, \#\#ati, \#\#tis}, neither of which are semantically revealing for our purposes.
For this reason, work based on WordPiece but with highly specialised corpora should always at least include a consideration of retraining the tokeniser and expanding the maximum vocabulary size.

\subsection{Model training}
\label{sub:model_training}

Model training was carried out over the course of a single day on a high-performance cloud appliance with 24 Intel Xeon E5-2690 CPUs at 2.60GHz and 440GB onboard RAM, equipped with four Nvidia Tesla V100-PCIe GPUs of 16GB each.
The model was trained with a batch size of 64 at an initial learning rate of $2 \times 10^{-5}$.
Hyperparameters for the Adam optimiser were set at $\beta_1 = 0.9$, $\beta_2 = 0.999$ and $\hat{\epsilon} = 10^{-8}$.
Training duration was set for 5 epochs, comprising 99,255 training steps, with periodic evaluation over the test set and checkpointing at 5,000-step increments.
The final model was drawn from the best-performing checkpoint, as defined by the highest $F_1$ score.

The total training duration comprised 21.82 hours, which included periodic checkpointing and backup.
Using energy consumption data and regionally correlated carbon intensity, the total CO$_2$ emission equivalent was estimated using \texttt{codecarbon} 2.3.3 at 6.55 kg CO$_2$eq.

\subsection{Evaluation}

The trained model was evaluated on a previously-unseen evaluation set generated according to the process described in Subsection~\ref{sub:source_data}, constituting 15\% of the entire sample or X records.
Evaluation metrics were obtained for the overall sample, as well as class-wise precision, recall and $F_1$.
Sensitivity and specificity analyses were carried out using \texttt{scikit-learn} v1.4.0.\cite{Pedregosa2011Scikit-learn:Python}

\section{Results}
\label{sec:results}

The best model performance on the test set showed an $F_1$ value of 0.88, superior -- albeit only to a small extent -- to the best comparator candidate models, with a precision of 0.74 and recall of 0.67. 
On the validation set, which was not used during training, the model yielded a precision of 0.84 and recall of 0.86, for an $F_1$ value of 0.85. 
Due to the significant class imbalance of the source data set, the class-wise results show an unsurprising corresponding imbalance.
On a class-by-class level, the model was strongest at predicting reports with no events ($F_1$ of 0.93), which dominated the training set. 
Predictions of singular events (e.g. hospitalisation or mortality on its own) was typically stronger than the predictive value for combinations of events, with the exception of ER attendance ($F_1$ of 0.56, \textit{vis-a-vis} 0.66 for hospitalisation and 0.76 for mortality alone).
Of the individual classes, the model performed least favourably on identifying hospitalisation events followed by mortality, but not including ER attendance ($F_1$ of 0.43). 

On the level of individual events rather than combinations thereof, the model once again performed best for no events ($F_1$ of 0.93), followed by mortality ($F_1$ of 0.83), hospitalisation ($F_1$ of 0.77) and, finally, ER attendance ($F_1$ of 0.60).
Of combinations of two or more events, ER attendance followed by hospital admission was the only one that was accurately predicted in a significant number.
It may be hypothesised that because ER attendance and hospital admission are not entirely conceptually distinct, and may -- especially by lay reporters -- be subject to some conceptual commingling. 
\texttt{Admitted to [the] ER} is a phrase that occurs in a number of reports, which of course is difficult to differentiate from admission \textit{from} the ER (or, indeed, no admission at all but merely care at the ER).

Overall, the model performance reflected the intrinsic strengths and weaknesses of the source data set.
The vast class imbalance, skewing towards no-event reports, inherently conditions any model trained on VAERS towards stronger performance in that respect.
This result highlights the importance of enhancing models on a wide range of uncommon manifestations, and supports the case for further developing such domain-specific LLMs with more data from uncommon manifestations of ADRs.

\begin{figure}
\centering
\includegraphics[width=\columnwidth]{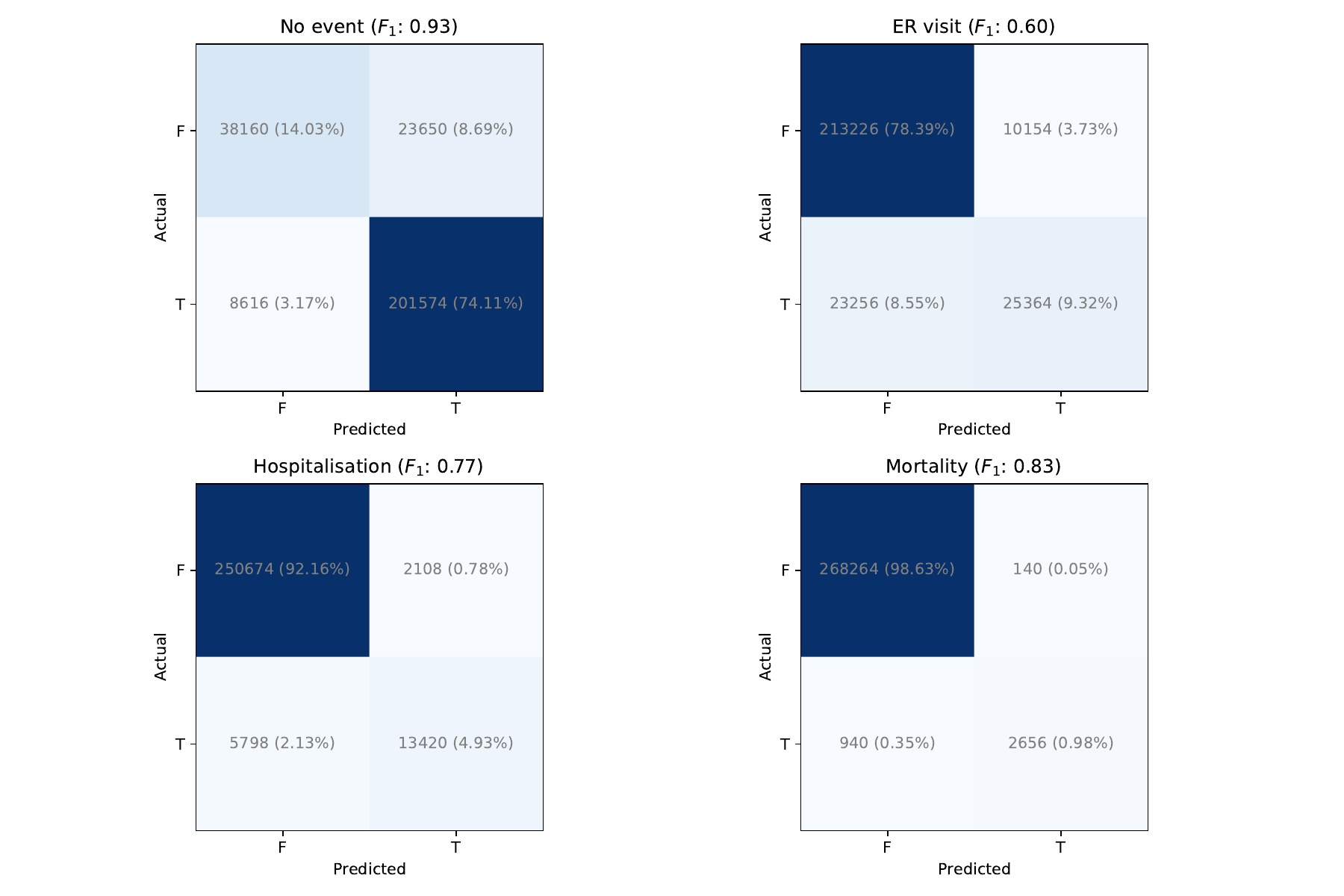}
\caption{Confusion matrices for the three events under consideration.}
\label{fig:confusion_matrices}
\end{figure}

\begin{figure*}
\centering
\includegraphics[width=16.5cm]{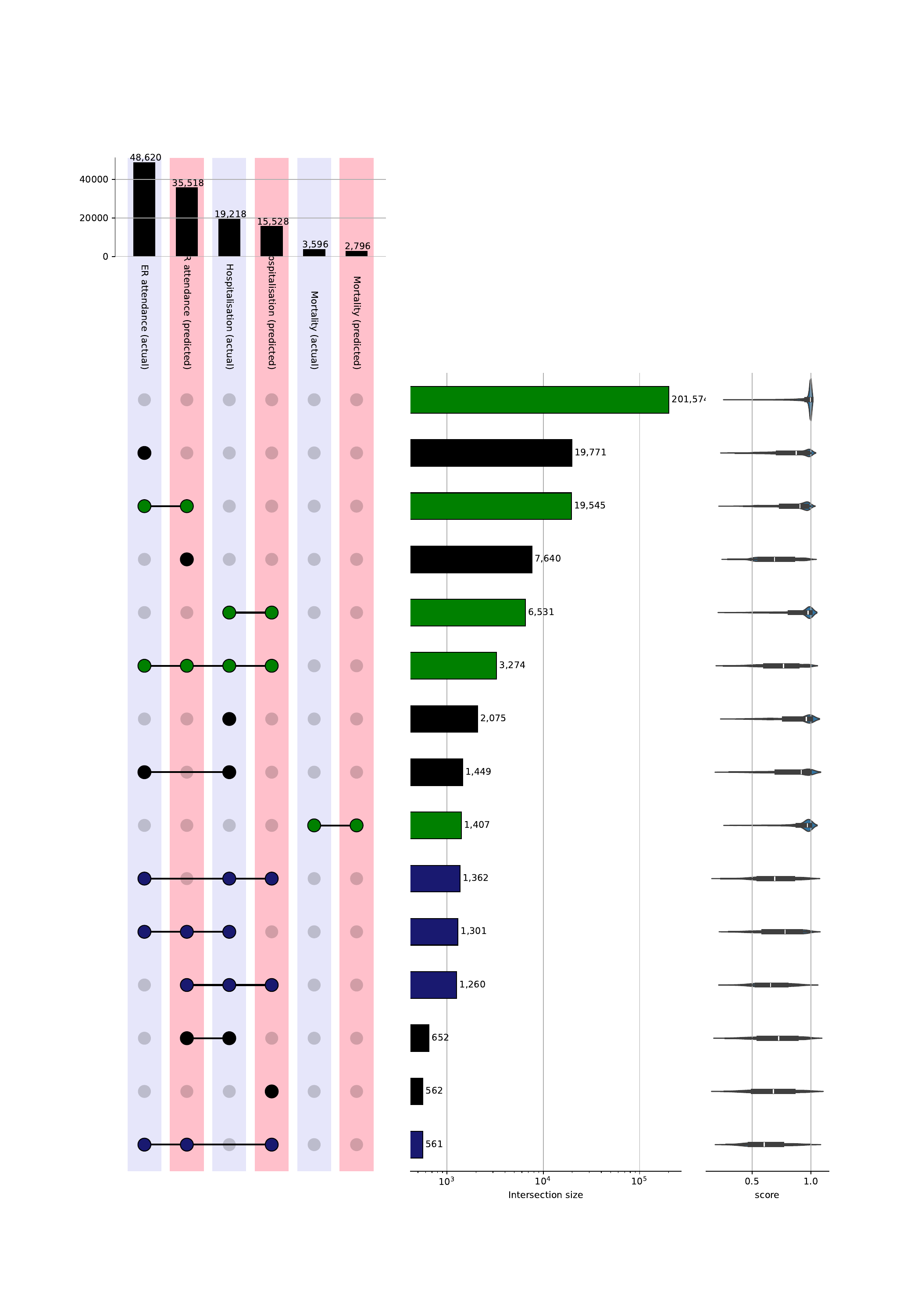}
\caption{Set combinations of predicted versus actual events. Correct predictions are displayed in green. Predictions that are partially correct, i.e. with respect to at least one event, are displayed in blue.}
\label{fig:upset_plot}
\end{figure*}

\begin{table}
\renewcommand{\arraystretch}{1.3}
\caption{Class-wise metrics for the final trained model.}
\label{tab:eval_metrics}
\centering
\begin{tabular}{l|ccc|c}

\hline

\bfseries Class & \bfseries Precision & \bfseries Recall & \bfseries $F_1$ & \bfseries Support \\

\hline \hline 

No event & 0.89 & 0.96 & 0.93 & 210,190 \\
ER attendance & 0.67 & 0.49 & 0.56 & 40,228 \\
Hospitalisation & 0.71 & 0.62 & 0.66 & 10,566 \\
Mortality & 0.81 & 0.72 & 0.76 & 1,947 \\
\makecell[l]{ER attendance + \\hospitalisation} & 0.58 & 0.44 & 0.50 & 7,420 \\
\makecell[l]{ER attendance + \\mortality} & 0.82 & 0.56 & 0.66 & 417 \\
\makecell[l]{Hospitalisation + \\mortality} & 0.59 & 0.34 & 0.43 & 677 \\
\makecell[l]{ER attendance + \\hospitalisation + mortality} & 0.67 & 0.47 & 0.55 & 555 \\
\hline
\end{tabular}
\end{table}

\section{Discussion}
\label{sec:discussion}

LLMs, and the computational infrastructure to train and fine-tune them at affordable costs and reasonable speeds, are rapidly attaining a point of maturity.
In this paper, we disclose the conception, design, development, training and evaluation of the first LLM trained on -- and for -- the pharmacovigilance domain.
From the translational perspective, we see three major contributions in our work.

First, it lends further support to the tenor of \citeauthor{Gururangan2020DontTasks}'s \citeyear{Gururangan2020DontTasks} paper, which argued for the value of domain-specific training of LLMs.\cite{Gururangan2020DontTasks}
Even though the overall impact of training the model on the entire training set (75\% of the entire VAERS corpus) has only created approx. 1.5\% uplift in terms of the $F_1$ score when compared to training the base model on only 10\% of the sample, such incremental gains are meaningful when one considers the absolute number of instances: this uplift would equate to roughly 15,000 more correctly classified records. 

Second, it highlights the value of subdomain-based training, but also of its limitations.
In particular, the fact that a general medical model (and even a general scientific model) proved superior to \citeauthor{Alsentzer2019PubliclyEmbeddings}'s specifically clinical model highlights the strong context dependence of subdomain models.
ADR reports are, of course, not the same as the clinical notes from MIMIC-III that \citeauthor{Alsentzer2019PubliclyEmbeddings}'s model was trained on: in particular, the MIMIC-III data set is derived from critical care medicine, whereas in our dataset, almost three in four records involved neither an ER/ED presentation, hospital admission or mortality event. 
The MIMIC-III data set is also entirely a product of clinicians, whereas our data set includes reports directly from patients and other laypersons, as well as reports made to the manufacturer over the telephone and submitted by the manufacturer to VAERS.
In language, context is everything, and the context-specificity of subdomain-specific models limits their utility (and considerable power) to their respective subdomain.
In that sense, such subdomain-specific models benefit, and suffer, from their specialisation: they perform better in their subdomain, but generalise less well.
The relative ease and affordability of training such models means that small, easy-to-train, low-parameter complexity but strongly domain-specific models remain a viable option for language tasks over corpora of highly specialised language.

Third, our work substantiates the continuing viability of low-impact training for developing such specialised models.
At 126 million parameters, our final model would be considered small in scale compared to the prevailing scale of SOTA models with billions to tens-of-billions of parameters.
However, the gains of increasing model size for highly specific models are rapidly diminishing as size increases.
This is countervailed by an exponentially increased training cost and training time.
The costs of training LLMs, from the immediate economic costs to the environmental burdens that arise from CO$_2$ emissions attributable to the not insignificant energy consumption of this process, have been widely documented.\cite{Bender2021OnBig}
In the field of global and public health, of which pharmacovigilance forms a part, our responsibilities of good stewardship of resources and environmental justice must be kept in mind when considering the right size of model to use.
Equally, smaller models are more accessible, allowing users in resource-constrained settings to benefit from inference and fine-tuning at an affordable cost.
The global proliferation of LLMs has created an interdependent ecosystem of models 'standing on the shoulders of giants', and creating models that are accessible to this process is thus becoming both a prudential and a moral imperative.

Of course, DAEDRA only explores one aspect of pharmacovigilance -- namely, certain outcomes, selected for their prevalence and their regulatory importance. 
Another limitation of this study is that its source data set comes entirely from one type of pharmaceutical -- namely vaccines --, and reports may thus not generalise to pharmacovigilance contexts of other preparations that present different issues (for instance, gastrointestinal distress -- one of the most frequently reported ADRs for non-vaccine pharmaceuticals -- is relatively uncommon in vaccine pharmacovigilance, while injection site reactions are equally uncommon outside the vaccine pharmacovigilance domain where oral preparations tend to dominate).
The study's source data was also limited to pharmacovigilance reports in English made in the United States.
Future research may fruitfully explore if expanding the purview of source data selection to non-vaccine products (e.g. by integrating FAERS data, which covers non-vaccine reports made to the FDA), data from passive reporting schemes in other countries that reflect different demographics (e.g. the UK's Yellow Card scheme or the European Medicines Agency's EudraVigilance service) or in other languages (e.g. domestic pharmacovigilance authorities' datasets from across the world) may bring new insights to our ability to leverage LLMs for reasoning about ADRs from passive reporting frameworks. 

\subsection*{Data availability}

The final matched dataset is available under \href{https://doi.org/10.57967/hf/1706}{\texttt{doi:10.57967/hf/1706}}. The final trained model is available under \href{https://doi.org/10.57967/hf/1757}{\texttt{doi:10.57967/hf/1757}}. All code and other supplementary materials that are relevant to this article are available on GitHub under \href{https://github.com/chrisvoncsefalvay/daedra}{\texttt{chrisvoncsefalvay/daedra}}.

\subsection*{Informed consent statement}
No IRB approval was sought because the research did not concern human subjects or identifiable patient data. All source data were publicly available from the CDC/FDA VAERS website.

\subsection*{Conflicts of interest}
CvC is a consultant to a company that may be affected by the research reported in this paper. The company had no influence upon the formulation, analysis or publication of this manuscript.

\bibliographystyle{IEEEtranSN}
\bibliography{references}

\end{document}